\documentclass[journal]{IEEEtai}

\usepackage[colorlinks,urlcolor=blue,linkcolor=blue,citecolor=blue]{hyperref}

\usepackage{color,array, makecell}

\usepackage{graphicx}
\usepackage{csquotes}% advanced facilities for inline and 

\usepackage{float}
\usepackage{hyperref}
\usepackage{eqparbox}
\usepackage{subfigure}
\usepackage{flushend}
\usepackage{cite}
\usepackage{tablefootnote}
\usepackage{mdwmath}
\usepackage{amssymb}
\usepackage{amsmath}
\usepackage{mdwtab}
\usepackage{url}
\usepackage{xcolor}
\usepackage{comment}
\usepackage{footnote}
\usepackage{multirow}
\usepackage{graphicx}
\usepackage{caption}

\begin{document}

\title{{\color{black}Multi-objective Representation for Numbers in Clinical Narratives: A CamemBERT-Bio-Based Alternative to Large-Scale LLMs}}

\author{Boammani Aser Lompo, Thanh-Dung Le,~\IEEEmembership{Member,~IEEE,} \\
	Philippe Jouvet M.D., Ph.D., and Rita Noumeir Ph.D.,~\IEEEmembership{Member,~IEEE}

	\thanks{This work was supported in part by the Natural Sciences and Engineering Research Council (NSERC), in part by the Institut de Valorisation des données de l’Université de Montréal (IVADO), and in part by the Fonds de la recherche en sante du Quebec (FRQS).}
	
	%}
\thanks{Boammani Aser Lompo is with the Biomedical Information Processing Lab, \'{E}cole de Technologie Sup\'{e}rieure, University of Qu\'{e}bec, Canada (Email: boammani.lompo.1@ens.etsmtl.ca)}

\thanks{Thanh-Dung Le is with the Biomedical Information Processing Lab, \'{E}cole de Technologie Sup\'{e}rieure, University of Qu\'{e}bec,  Montr\'{e}al, Qu\'{e}bec, Canada (Email: thanh-dung.le.1@ens.etsmtl.ca).}

\thanks{Philippe Jouvet is with the CHU Sainte-Justine Research Center, CHU Sainte-Justine Hospital, University of Montreal, Montr\'{e}al, Qu\'{e}bec, Canada.}

\thanks{Rita Noumeir is with the Biomedical Information Processing Lab, \'{E}cole de Technologie Sup\'{e}rieure, University of Qu\'{e}bec,  Montr\'{e}al, Qu\'{e}bec, Canada.}

} \author{Boammani Aser Lompo, Thanh-Dung Le,~\IEEEmembership{Member,~IEEE,} \\
Philippe Jouvet M.D., Ph.D., and Rita Noumeir Ph.D.,~\IEEEmembership{Member,~IEEE}

\thanks{This work was supported in part by the Natural Sciences and Engineering Research Council (NSERC), in part by the Institut de Valorisation des données de l’Université de Montréal (IVADO), and in part by the Fonds de la recherche en sante du Quebec (FRQS).}

%}
\thanks{Boammani Aser Lompo is with the Biomedical Information Processing Lab, \'{E}cole de Technologie Sup\'{e}rieure, University of Qu\'{e}bec, Canada (Email: boammani.lompo.1@ens.etsmtl.ca)}

\thanks{Thanh-Dung Le is with the Biomedical Information Processing Lab, \'{E}cole de Technologie Sup\'{e}rieure, University of Qu\'{e}bec,  Montr\'{e}al, Qu\'{e}bec, Canada, and also is with the Interdisciplinary Centre for Security, Reliability, and Trust (SnT), University of Luxembourg, Luxembourg.(Email: thanh-dung.le@uni.lu).}

\thanks{Philippe Jouvet is with the CHU Sainte-Justine Research Center, CHU Sainte-Justine Hospital, University of Montreal, Montr\'{e}al, Qu\'{e}bec, Canada.}

\thanks{Rita Noumeir is with the Biomedical Information Processing Lab, \'{E}cole de Technologie Sup\'{e}rieure, University of Qu\'{e}bec,  Montr\'{e}al, Qu\'{e}bec, Canada.}

} 

\maketitle

\begin{abstract}
The processing of numerical values is a rapidly developing area in the field of Language Models (LLMs). Despite numerous advancements achieved by previous research, significant challenges persist, particularly within the healthcare domain. This paper investigates the limitations of Transformers Models in understanding numerical values. \textit{Objective:} The aim of this research is to categorize numerical values extracted from medical documents into eight specific physiological categories using CamemBERT-bio. \textit{Methods:} In a context where scalable methods and Large Language Models (LLMs) are emphasized, we explore the option of lifting the limitations of transformer-based models. We examine two strategies: fine-tuning CamemBERT-bio on a small medical dataset with the integration of Label Embedding for Self-Attention (LESA), and combining LESA with additional enhancement techniques such as Xval. Given that CamemBERT-bio is already pre-trained on a large medical dataset, the first approach aims to update its encoder with newly added label embeddings technique, while the second approach seeks to develop multiple representations of numbers (contextual and magnitude-based) to achieve more robust number embeddings. \textit{Results:} As anticipated, fine-tuning the standard CamemBERT-bio on our small medical dataset did not improve F1 scores. However, significant improvements were observed with CamemBERT-bio + LESA, resulting in an over 13\% increase. Similar enhancements were noted when combining LESA with Xval, outperforming conventional methods and giving comparable results to those of GPT-4 \textit{Conclusions and Novelty:} This study introduces two innovative techniques for handling numerical data, which are also applicable to other modalities. We illustrate how these techniques can improve the performance of Transformer-based models, achieving more reliable classification results even with small datasets.
\end{abstract}

\IEEEpeerreviewmaketitle
\begin{IEEEkeywords}
Clinical Natural Language Processing, Numerical values classification, Xval, LESA
\end{IEEEkeywords}

\textbf{\textit{Clinical and Translational Impact Statement---} The use of CamemBERT-bio to numerical value classification holds significant promise, particularly given the limited availability of clinical data. By developing multiple representation-based embeddings for numbers, we enhance our models robustness and generalizability. This approach enables the creation of diagnostic algorithms from unstructured medical notes, as our model provides the necessary structure. The methodology can be extended to various other NLP challenges and modalities, leveraging the core idea of constructing embeddings based on multiple data representations. Additionally, this work addresses the common healthcare issue of handling small and imbalanced datasets.}

\section{Introduction}

\IEEEPARstart{A}{} major challenge in processing medical texts is understanding numerical data. Numbers are ubiquitous in medical documents, appearing in various forms such as physiological measurements, DNA coding sequences, event occurrences (e.g., number of births, cardiac failures), durations (e.g., pregnancy length, medication duration), and medical protocol codes (e.g., G1P2 standing for Gravida 1 Para 2). In the context of Clinical Decision Support Systems (CDSS), it is crucial to recognize and interpret the meaning of these numbers within their specific context to determine if they indicate a potential problem. For example, in the text "Patient agé de 12 ans, rythme cardiaque de 130 bpm," it is essential to identify "12" as the age of the patient and "130" as the heart rate, and then understand that a heart rate of "130" is problematic for a 12-year-old patient.

Most machine learning efforts in the medical domain must contend with small and imbalanced datasets due to the high cost of data annotation in healthcare. Additionally, medical notes are often unstructured, filled with medical jargon, and sometimes contain incomplete sentences. This necessitates handling shorthand notations, typographical errors, and various forms of textual noise commonly found in medical documentation, making systematic approaches like case-by-case algorithms ineffective. In addition, numerical values in medical datasets typically span a wide range. For example, the dataset provided by Centre Hospitalier Universitaire Sainte-Justine (CHUSJ) for our study includes numerical values ranging from $10^{-4}$ to $10^5$. Therefore, careful gradient management is essential, as a single large outlier can negatively impact the overall performance of the model. 

This paper contributes to a broader project aimed at leverage Language Models for diagnosing cardiac failure. According to the National Heart, Lung, and Blood Institute (2023), cardiac failure occurs when the heart is unable to pump sufficient blood to meet the body's needs, either due to inadequate filling or diminished pumping capacity. Cardiac failure presents through a range of physiological indicators, including blood pressure, ventricular gradients, and ejection fractions. Our research focuses on classifying numerical values extracted from medical notes into one of eight specific physiological categories using CamemBERT-bio \cite{touchent2023camembert}.

\subsection{Goal statement} 
Our objective is to train CamemBERT-bio on a limited and uneven dataset with the aim of accomplishing the following outcome:

\noindent \textbf{Input medical note}:
\begin{displayquote}
\textit{``Norah, BB né à 37+6 par AVS, APGAR 8-8-8. Dx Anténatale: d-TGV avec septum intact, sat 50-65\% à la naissance."}
\end{displayquote}
\textbf{Expected output}:

\begin{table}[H]
%\caption{Table captions should be placed above tables.}
\label{tab1}
{
\centering
\begin{tabular}{|l|l|l|}
\hline
Value &  Attributes & Unit\\
\hline
$37+6$ & Divers (âge du patient) & semaines\\
$8-8-8$ & score APGAR & \\
$50-65$ & saturation en oxygène & {\bfseries$\%$}\\
\hline
\end{tabular}\par
}
\end{table}

\subsection{Contribution} 
In this work, we outline the subsequent contributions:
\begin{itemize}
\item We demonstrate that CamemBERT-bio + LESA as introduced in \cite{lompo2025mediumsizedtransformersmodelsrelevant} can be further enhanced through language modeling prefinetuning. This prefinetuning does not improve CamemBERT-bio alone, indicating that integrating the Label Embedding for Self-Attention (LESA) technique necessitates pretraining to adjust the model's parameters. Prefinetuning on a small medical dataset significantly improves the F1 score for CamemBERT-bio + LESA.
\item We develop a new model by extending finetuned CamemBERT-bio + LESA with the Xval architecture \cite{golkar2023xval}. The main idea of Xval is to incorporate the magnitude of numbers into their embeddings before the attention layers. This new model provides a multi-objective representation of numbers, enhancing generalizability and robustness.
\item We assess and juxtapose our tailored model against conventional counterparts {\color{black}including LLMs such as GPT-4}. Findings demonstrate that our model surpasses standard approaches and adeptly manages datasets characterized by limited size and imbalance. {\color{black}Additionally, our model delivers performance on par with GPT-4 despite the significant disparity in parameter size.}
\end{itemize}

\section{\label{related} Related works}

Processing numerical values with LLMs is a highly active research area. While \cite{wallace2019nlp} claimed that most word embedding models could capture numeracy, several limitations persist. Various approaches have been developed to handle numerical values. \cite{cui2019regular} tackled the issue of number classification by employing a rules-based method utilizing regular expressions. Nevertheless, the success of this approach is largely contingent on the clarity of the text. It depends significantly on consistent abbreviations and the minimal presence of irrelevant textual noise, which can disrupt the regular expression generator's accuracy. \cite{chen2023improving} proposed a question-answer pre-training task aimed at improving number understanding. Other studies have explored the potential of different number tokenization methods, such as digit-by-digit, scientific notation, and customized notations \cite{charton2021linear}. Some previous studies have tackled the challenge with a number-agnostic approach, replacing numbers with placeholder keywords and scaling their embeddings to enforce context-centric encoding \cite{golkar2023xval, Loukas_2022,  lompo2025mediumsizedtransformersmodelsrelevant}. In their work, \cite{lompo2025mediumsizedtransformersmodelsrelevant} introduced Label Embedding for Self-Attention (LESA), which incorporates class descriptions directly into the initial feature extraction phase to help the feature extractor focus on relevant features for classification. LESA is similar to EmbedNum \cite{Nguyen2018EmbNumSL}, which classifies numbers by comparing their embeddings to those of potential categories. However, LESA differs by using similarity exclusively to construct the number's embedding rather than for the classification process itself. {\color{black} In the current state of the art, advanced LLMs such as LLAMA, GPT-4, Claude are engineered to interpret and adjust to diverse contexts autonomously, making it possible to handle numerical values without requiring any preprocessing.}

{\color{black}
Recent studies have highlighted several limitations of large language models (LLMs) despite their strong performance across various tasks. While increasing model size has generally led to improved results, some research suggests that LLMs continue to struggle with complex semantic understanding \cite{cheng2024potential}. For instance, \cite{shah2024accuracy} found that GPT-3.5 demonstrated high efficiency in extracting critical information from medical notes, but this came at the cost of accuracy, as the model remained prone to errors involving abbreviations and misinterpretations. Similarly, \cite{ullah2024challenges} identified significant issues in contextual understanding, interpretability, and bias in training data when evaluating GPT-4 for medical applications. Their findings indicate that these challenges stem from the model's limited grasp of medical concepts and the insufficient representation of real-world clinical data during training, which could lead to errors and disparities in medical diagnoses. Additionally, the deployment of large-scale LLMs such as GPT-4 raises critical concerns regarding patient data privacy and security, as these models inherently retain vast amounts of training data within their parameters. Given these challenges and the high computational costs associated with LLMs, we prioritize the enhancement of smaller-scale language models, such as CamemBERT-bio, which provide a more adaptable and resource-efficient alternative for medical NLP applications.}.

Another significant challenge is handling small and imbalanced datasets. While transformer-based architectures generally perform well with large datasets, research by \cite{Li2018, ezen2020comparison, mascio2020comparative} indicates that they may not outperform traditional methods, such as LSTM, when dealing with smaller datasets of clinical notes. A common technique to enhance Language Models performance on small datasets is Knowledge Distillation (KD) \cite{Bucila2006ModelC}. Knowledge Distillation (KD) is based on the principle that a model with fewer parameters is less prone to overfitting. This technique entails training a smaller neural network to replicate the performance of a larger one, using a teacher-student framework. This methodology is the key idea of like DistilBERT \cite{Sanh2019} and its French counterpart, DistillCamemBERT \cite{delestre2022distilcamembert}. Another approach to handling limited data is to create multiple representations for words, leveraging diverse angles of approach to achieve more a robust and generalizable modeling. Typically, this involves constructing an objective loss as the weighted sum of various losses, each specialized in different aspects of the studied modality. However, this approach raises the critical issue of appropriately weighting each loss to avoid bias. In \cite{kendall2018multi}, the authors propose an automatic method for weighting multiple losses by optimizing the log-likelihood of the model predictions. Although their approach was applied to images, we found their results relevant and potentially beneficial for our case.

{\color{black}
In summary, handling numerical values in NLP remains a challenging and evolving research area. While various approaches, such as rule-based methods, number tokenization techniques, and number-agnostic representations, have been explored, recent advancements in LLMs have enabled more autonomous and context-aware processing of numerical data without requiring extensive preprocessing. However, despite their impressive performance, LLMs still face notable limitations, including challenges in semantic understanding, susceptibility to errors in medical contexts, interpretability issues, and concerns regarding data privacy. Given these challenges, smaller-scale language models, such as CamemBERT-bio, offer a viable alternative by balancing performance, efficiency, and adaptability in specialized domains. Additionally, addressing the challenges of small and imbalanced datasets remains crucial, with techniques like KD and multi-loss optimization proposed to enhance model performance in low-data scenarios.}

\section{\label{task and dataset}The task and the dataset}

The dataset utilized in this study was supplied by the Pediatric Intensive Care Unit at CHU Sainte-Justine (CHUSJ), and includes children under the age of 18. It comprises 26,671 unannotated medical notes from \textbf{2,514 de-identified patients}. Of these notes, only 1,072 samples (approximately 30,000 tokens) are annotated. All the details regarding the dataset are provided in \cite{lompo2025mediumsizedtransformersmodelsrelevant}.

\begin{figure}
\centering
\begin{subfigure}
  \centering
  \includegraphics[width=.7\linewidth]{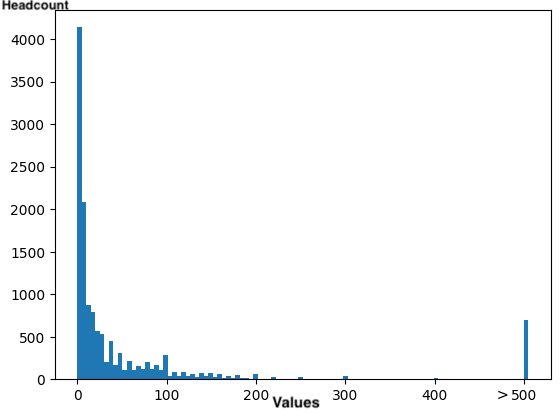}
  %\caption{}
  %\label{without lesa}
\end{subfigure}%
\begin{subfigure}
  \centering
  \includegraphics[width=.7\linewidth]{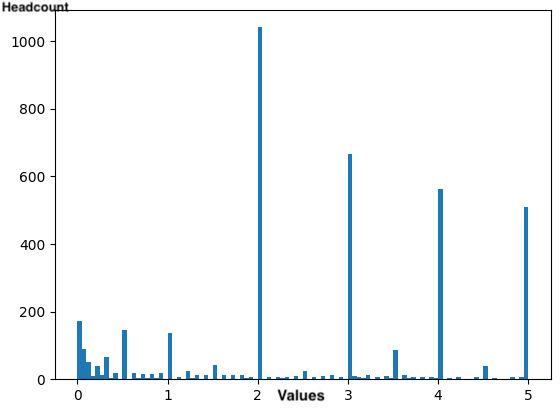}
  %\caption{}
  %\label{with lesa}
\end{subfigure}
\caption[Numbers distribution]{(a) Overlook to the whole numbers distribution (b) A focus on the small numbers distribution. Non integers numbers are effectively represented in the dataset.}
\label{fig:number distribution}
\end{figure}

The number distribution is displayed in Figure ~\ref{fig:number distribution}. All the numbers beyond $500$ are regrouped because they are scattered from $500$ to $35000$.

\section{A few limitations of end-to-end learning for Transformers Models in medical datasets}
{\color{black} Most pretrained Transformers Foundational Models such as GPT-4, LLAMA and BERT are pretrained in a end-to-end paradigm.} End-to-end learning involves training a model to perform a task from raw input to final output, without intermediate steps or feature engineering. While this is the standard approach for training most Transformers, our work faces the significant limitation of working with small and imbalanced datasets, unlike many state-of-the-art approaches mentioned earlier. This section explores other major limitations of end-to-end training for Transformers Models in the medical domain.

\subsection{\label{tokenization problem}The Limitations of Tokenization}

A fundamental aspect of recent transformer-based models revolves around the introduction of highly efficient tokenization techniques, notably exemplified by WordPiece \cite{wu2016google} and SentencePiece \cite{kudo-richardson-2018-sentencepiece}. These tokenizer variants share a common trait: the capability to deconstruct intricate or infrequent words into subsequences comprising their constituent characters, an important ability in managing Out-of-vocabulary (OOV) words. Let's consider the word \textbf{``working"}, which undergoes tokenization into \textbf{``work"} + \textbf{``ing"}. This resultant sequence forms a good starting point to model the word, as \textbf{``work"} and \textbf{``ing"} independently hold semantic significance and individual meaning. However, as mentionned by \cite{Loukas_2022}, we claim that this approach exhibits limitations when dealing with numerical values. For instance, consider the instance of \textbf{``123.7g/L"}, which would be disassembled into \textbf{``12"} + \textbf{``3"} + . + \textbf{``7"} + \textbf{``g/L"}. This segmentation raises several issues. Predominantly, in many classification tasks, only the first token of a segmented word is subject to classification. Only \textbf{``12"} would be classified in our example. However, \textbf{``12"} is markedly ambiguous, vastly deviating from the initial term's inherent meaning. Furthermore, this method of breaking down numerical tokens might fail to capture their magnitudes. The difference in magnitude between \textbf{``12"} and \textbf{``123.7"} is substancial. While the self-attention mechanism and index position embedding hold the potential to discern this magnitude discrepancy, we claim that these mechanisms alone are insufficient. 

\begin{figure}
\centering
\includegraphics[width=1\linewidth]{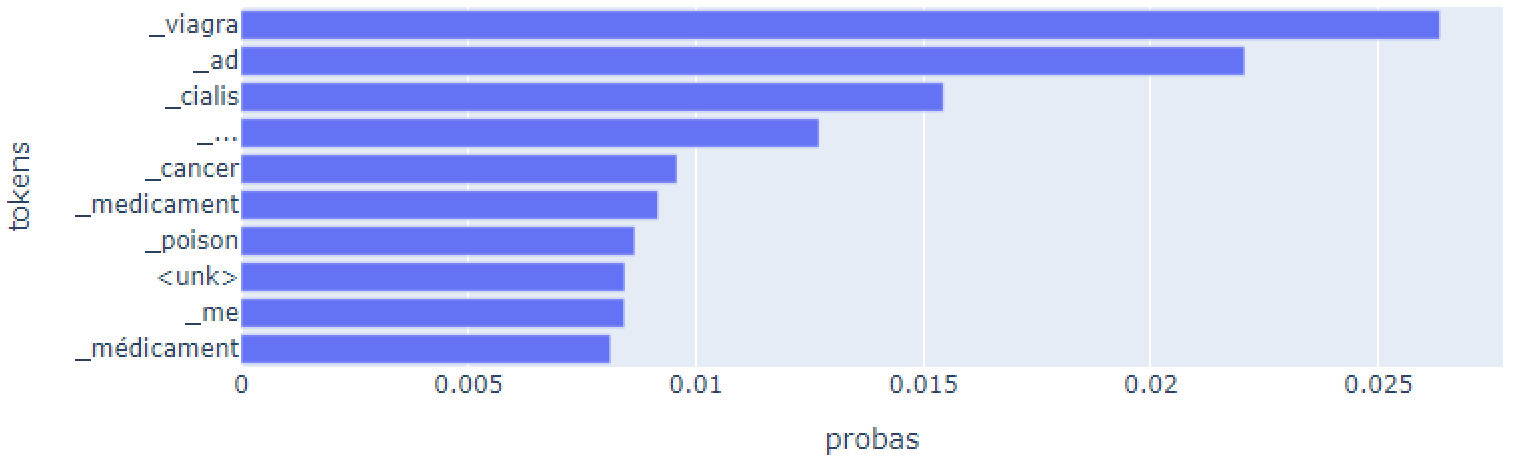}
  \caption{Output of a completion task by CamemBERT-bio. The input sentence is \textit{``Patient en détresse respiratoire, gradient VG-VD ad $\langle$ mask $\rangle$ mmgh."}. The figure contains the possible words to fill the mask with their corresponding probabilities.}
  \label{completion with camembert bio}
\end{figure}

To illustrate this point, let's examine CamemBERT-bio's performance in a task focused on completing sentences that contain numerical values (Figure ~\ref{completion with camembert bio}). A word completion task involves taking a sentence and replacing one or more words with the token $\langle$mask$\rangle$. The model is then asked to predict the missing words. The input sentence, \textit{``Patient en détresse respiratoire, gradient VG-VD ad $\langle$ mask $\rangle$ mmgh."}, was crafted to align with the language style of our target dataset. It becomes evident from this illustration that the model struggles to provide meaningful numerical predictions as the predicted tokens are not even numbers. 

\subsection{\label{structure issue}The Structure Issue}

Some earlier research focused on studying transformer-based models to understand how they function and evaluate their ability to recognize different syntactic relations. \cite{clark2019does} revealed that, surprisingly, without explicit supervision on syntax and co-reference,  certain heads in the BERT model specialize in grammatical tasks such as finding direct objects of verbs, determiners of nouns, objects of prepositions, and objects of possessive pronouns with more than 75\% accuracy. This reveals that the structure of sentences matters for these models. However, our medical notes landscape stands apart in this regard. A significant proportion of sentences within these notes lack well-defined structures, with instances wherein clear subjects and verbs are even absent, as aptly exemplified by the following illustrations:

\begin{itemize}
    \item ``\textit{CARDIOMYOPATHIE  1) S'est présenté le 28/04 à l'hôpital - Admis ad 07/05 aux soins [...] Depuis, céphalée on/off, fatigue, essoufflement.[...] Vomissement le soirx 5 jours.\phantom{eeeee}    Pincement thoracique à 2-3 reprises depuis 1mois. l'a réveillé 1 fois.}"
    \item ``\textit{ATCDp: \#PNA droite à 1 mois de vie et en 2004 \phantom{eeeee}  Atrophie/asymétrie rénale 2aire   Vu en néphro: \phantom{eeeee}   Écho (11/2014): Rein Droit: 8.8 cm, contour supérieur légèrement irrégulier + vague zone kystique, de 5 mm, [...].}"
    \item \textit{``[...] FR 21 FC 100-110 FR 50 [...]."}
\end{itemize}

As noted in \cite{lompo2025mediumsizedtransformersmodelsrelevant}, the last excerpt illustrate the lack of contextual information in sentence structures. This excerpt reads a sequence of physiological indicators. The initial reference to \textit{FR} most likely indicates the respiratory rate (fréquence respiratoire) since a value of 21 is usually too low for the shortening fraction (fraction de raccourcissement). Conversely, the second occurrence of \textit{FR} refers to a value above 50, suggesting it pertains to the shortening fraction. In this case, it was the values themselves, rather than the context, that provided the necessary distinctions, highlighting a flaw in the sentence structure. This issue echoes the challenges of tokenization, emphasizing the need for a thorough understanding of the numerical values. {\color{black} However, it is worth mentioning that the impact of this issue is very limited on advanced LLMs such as LLAMA, GPT-4, Claude that are engineered to be able to work with data without requiring any preprocessing. Nevertheless, their ability in handling the lack of structure significantly on their prompts \cite{hu2024improving}}.

\section{\label{approach}Methodology}
The methodology adopted in this study involves developing a multi-objective representation of numbers to achieve a more generalizable and robust model \cite{collobert2008unified}. Our representations are obtained through the Masked Language Modelling (MLM) Loss and the Number Prediction Loss. The application of these two representations result in the two following models:

\begin{itemize}
    \item CamemBERT-bio + LESA prefinetuned via MLM on our small medical dataset
    \item CamemBERT-bio + LESA + Xval trained with both MLM loss and Number Prediction Loss.
\end{itemize}

We anticipate that this multi-objective training approach will produce number representations that incorporate both contextual and magnitude information.

\subsection{\label{mlm}Masked Language Modelling (MLM) objective with LESA}

We summarize the model description from \cite{lompo2025mediumsizedtransformersmodelsrelevant}. Let's consider a text $\boldsymbol{t} = [token_1, token_2, \cdots, token_L]$ of $L$ tokens, with initial embeddings $[\boldsymbol{x}_{token_1}, \cdots, \boldsymbol{x}_{token_L}]$. These initial embeddings, prefixed with the special token $[CLS]$, are fed into the transformer layers in the following manner:
\begin{equation}    
\boldsymbol{X} = [\boldsymbol{x}_{CLS}, \boldsymbol{x}_{token_1}, \cdots, \boldsymbol{x}_{token_L}] \in \mathbb{R}^{(L+1)\times D}
\end{equation}

\noindent where $D$ is the dimension of the embedding space. 

\noindent For every individual class, some descriptive keywords are selected, and their initial embeddings are computed and then averaged. This computation produces an embedding matrix, represented as $\boldsymbol{X}^{l}\in \mathbb{R}^{n\times D}$, where $n$ corresponds to the number of classes. This matrix includes unique label embeddings for each class. 

Subsequently, the authors compute the cross-attention between $\boldsymbol{X}^{l}$ and $\boldsymbol{X}$, $\boldsymbol{A}^{l} \in \mathbb{R}^{n\times (L+1)}$, and the self-attention of $\boldsymbol{X}$ $\textbf{\textrm{Self-attention}}\in \mathbb{R}^{(L+1)\times (L+1)}$.

Then, after computing the cosine similarity matrix $\textbf{\textrm{CoSim}} =  \textrm{norm}(\boldsymbol{A}^{l})^T \textrm{norm}(\boldsymbol{A}^{l})$, we therefore get the enhanced self-attention:
\begin{equation}
     \textbf{\textrm{New-Self-attention}} = \textbf{\textrm{Self-attention}} + \textbf{\textrm{CoSim}}
\end{equation}

\begin{figure}
\centering
\includegraphics[scale=0.46]{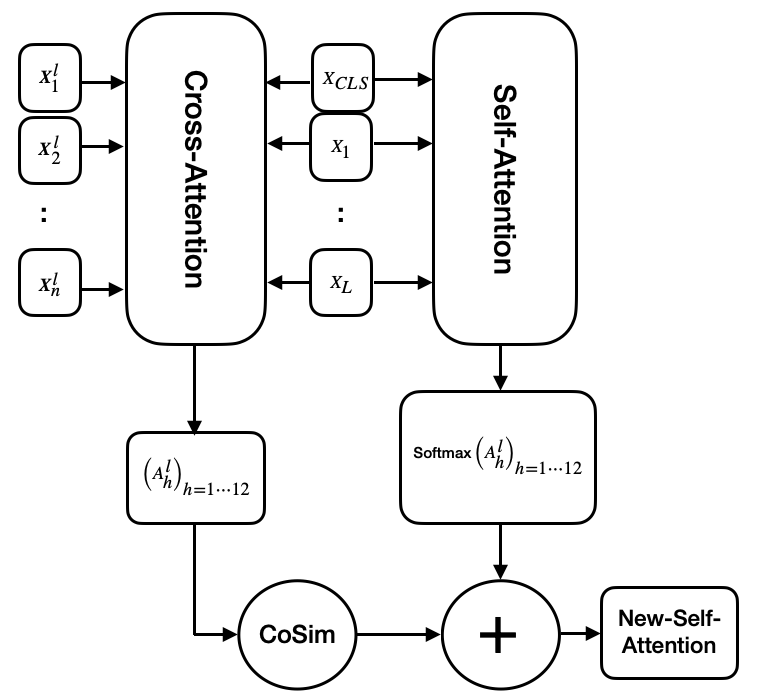}
  \caption{Excerpt from \cite{lompo2025mediumsizedtransformersmodelsrelevant}. An illustration of the Label Embedding for Self-Attention (LESA) is shown. The input to this Self-Attention layer consists of the token embeddings $[X_{CLS}, X_1, \cdots, X_L]$ and the keyword embeddings $[X^l_1, X^l_2, \cdots, X^l_n]$. This layer outputs the enhanced self-attention.}
  \label{fig:lesabert}
\end{figure}

In this work, we use this approach to address the limited size of our dataset as it reduces the risks of overfitting while making the model more robust. We then train the model via the Masked Language Modelling (MLM) task as a prefinetuning step. The MLM task involves randomly masking tokens and asking the model to predict the masked tokens, thereby training the model to build contextual representations of tokens.

\subsection{\label{xval} Number Prediction Objective}
This method consists in four steps as detailed in \cite{golkar2023xval}. Let's consider an input text $t$ containing both numbers and words. First, all the numbers are replaced by a placeholder keyword \textit{NUM}, forming a new input text $t_{masked}$. The initial numbers are saved for later. Then, we tokenize and embed $t_{masked}$ resulting in the matrix $h_{masked}$. Subsequently, all the vectors in $h_{masked}$ corresponding to the numbers previously replaced are multiplied by their corresponding value, resulting in a new matrix $h$. Finally, this embeddings matrix $h$ is fed to the encoder of our model. The pipeline is illustrated in Figure ~\ref{fig:diagram multi}.

By replacing all numbers by a keyword placeholder, we solve the tokenization limitation mentionned earlier. This loss objective aims at training the model to learn the magnitude of the numbers it encodes. The magnitude information is very important as it is a good discriminator between classification categories. For instance the numbers belonging to heart rate class are usually bigger than those belonging to APGAR class. 

\subsection{Multi Objective Loss}

Working with multiple objectives losses require to merge them into one single loss
$$L = \sum w_iL_i $$
where $w_i$ represents the weight of the loss $L_i$. {\color{black} In our case we have two losses $L_1$ which corresponds to the MLM objective and $L_2$ which corresponds to the Number Prediction Objective. Our training dataset $\mathcal{D}$ is made of couples $(x,y_1, y_2)$ where $x$ is a sequence of tokens with one token masked using the keyword \textit{Mask}, $y_1$ is the index of the masked token in the model vocabulary, and $y_2$ is the value of the masked token if it was a number. Otherwise $y_2$ is set to $1$. Figure ~\ref{fig:diagram multi} illustrates the inference process of the model. In this illustration, the first masked token is decoded as the standard word "Saturation," and the default number $1$ is assigned to it. The second masked token is decoded as the special token "NUM," and the number $65$ is predicted by the number prediction head.

\begin{figure}
\centering
\includegraphics[scale=0.335]{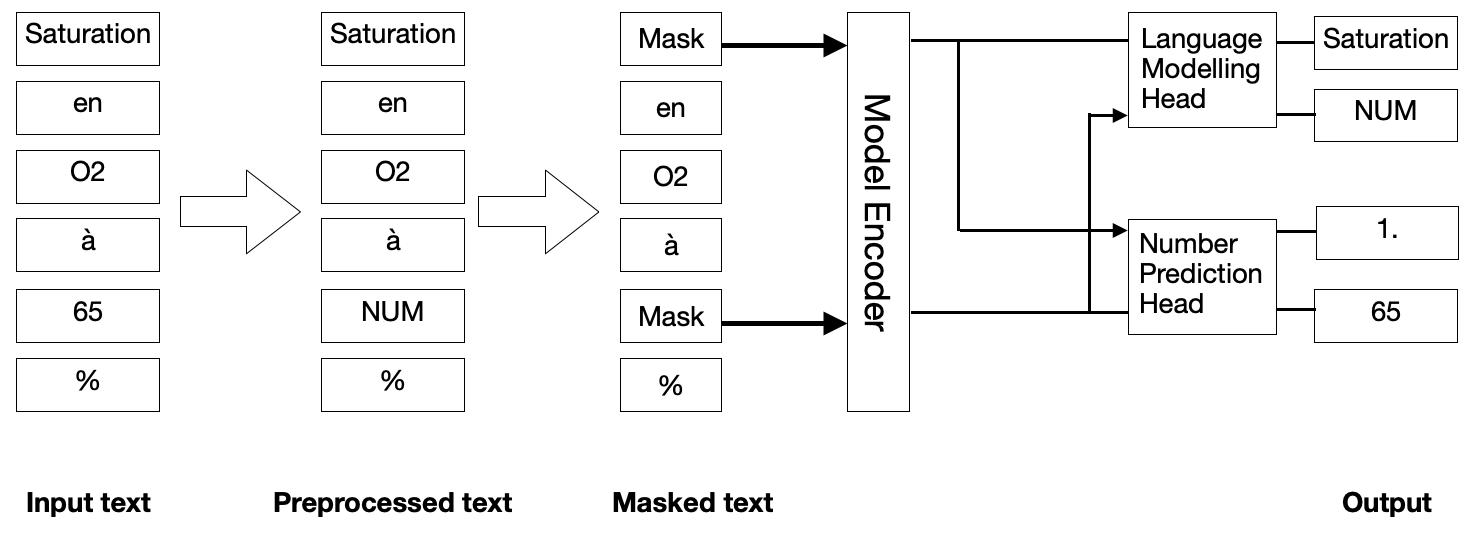}
  \caption{Brief illustration of the inference process using multiple representations}
  \label{fig:diagram multi}
\end{figure}

In \cite{kendall2018multi}, the authors propose an automatic way to compute each loss weight $w_i$. For the MLM objective, the predicted token distribution is usually given as a Softmax:

\begin{equation}
   p(z|f_1(x, \Theta), \sigma_2) = \textrm{Softmax}\left(\frac{1}{\sigma_1^2}f_1(x,\Theta)\right)_z
\end{equation}
where $x$ is the input sequence, $f_1(x,\Theta)$ is the vector of logits outputted by the Language Head, $\Theta$ is the model parameters, and $\sigma_1$ is artificially added as a \textit{temperature} parameter. Then, after a few approximations, they get to write the log likelihood as 

\begin{equation}
    \log p(z|f_1(x,\Theta)) \propto \frac{1}{\sigma_1^2}\log\textrm{Softmax}\left(f_1(x,\Theta)\right)_z  - \log\sigma_1
\end{equation}
Here, one may recognize the cross entropy loss with its weight $\frac{1}{\sigma_1^2}$ and a regularization term $\log\sigma_1$. Therefore we take $L_1$ as the cross entropy loss function and we define $w_1 = \frac{1}{\sigma_1^2}$.

For the number prediction task, the authors predict the masked number using a Gaussian distribution:

\begin{equation}
    p\left(s|f_2(x, \Theta)\right) = \mathcal{N}\left(f_2(x, \Theta)|\sigma_2^2\right)
\end{equation}
where $f_2(x,\Theta)$ is Number Prediction Head's output, and $\mathcal{N}$ refers to the Gaussian probability distribution function and $\sigma_2$ its standard deviation. This leads to the log likelihood

\begin{equation}
    \log p(s|f_1(x,\Theta)) \propto -\frac{1}{2\sigma_2^2}\left(s-f_2(x,\Theta)\right)^2 - \log\sigma_2
\end{equation}
Then one can recognize the square error $\left(s-f_2(x,\Theta)\right)^2$ with its weight $\frac{1}{2\sigma_2^2}$ and a regularization term $\log\sigma_2$. Therefore, we take $L_2$ as the square error loss and we define $w_2 = \frac{1}{2\sigma_2^2}$.

Finally, when we sum up everything, we get the following loss function:

\begin{equation}
\label{eq:multi}
\begin{split}
    L(x, y_1, y_2) &= - \frac{1}{\sigma_1^2}\log\textrm{Softmax}\left(f_1(x,\Theta)\right)_{y_1} + \frac{1}{2\sigma_2^2}\left( y_2-f_2(x,\Theta)\right)^2  \\
    &+ \log\sigma_1 + \log\sigma_2\\
    &= \frac{1}{\sigma_1^2}L_1(x, y_1) + \frac{1}{2\sigma_2^2}L_2(x, y_2)  + \log\sigma_1 + \log\sigma_2
\end{split}
\end{equation}

More precisely the global objective function is
$$\mathcal{L}(\Theta) = \mathbb{E}_{\mathcal{D}}\left[L(x, y_1, y_2)\right]$$
with respect to $\Theta, \sigma_1, \sigma_2$ \cite{kendall2018multi}. The first-order conditions with respect to $\sigma_1$ and $\sigma_2$, give

\begin{equation}
\label{eq:1}
\sigma_1^2 =  2\mathbb{E}_{\mathcal{D}}\left[L_1(x, y_1)\right]    
\end{equation}

\begin{equation}
\label{eq:2}
\sigma_2^2 =  \mathbb{E}_{\mathcal{D}}\left[L_2(x, y_2)\right]    
\end{equation}

Since the exact numbers within a medical note are often not predictable from the context, and given the wide range of values present in the dataset (Figure ~\ref{fig:number distribution}), it is natural to assume that $\mathbb{E}_{\mathcal{D}}\left[L_2(x, y_2)\right]$ is significantly larger than $\mathbb{E}_{\mathcal{D}}\left[L_1(x, y_1)\right]$. Therefore by Equations ~\ref{eq:1} and ~\ref{eq:2}, $w_2$ is significantly smaller than $w_1$. Our experiments confirms this assumption, placing greater emphasis on contextual representation over magnitude-based representation. As a result, we observe poor performance.

$\mathbb{E}_{\mathcal{D}}\left[L_2(x, y_2)\right]$ depends on the numbers distribution in $\mathcal{D}$ and more precisely on the range of the distribution. In order to lower the impact of this parameter we substituted $L_2$ with its log scaled version $\log (L_2)$. Additionnally, we removed the regularization terms $\log\sigma_1$ and $\log\sigma_2$ and we adopted $w_1 = w_2 = \frac{1}{2}$

\begin{equation}
\tilde{L}_2 (x, y_2) = \left( \log(y_2+1) - \log\left(f_2(x,\Theta)+1\right)\right)^2 
\end{equation}

We therefore get the new multi-objective loss
\begin{equation}
    \tilde{L}(x, y_1, y_2) = \frac{1}{2}L_1(x, y_1) + \frac{1}{2}\tilde{L}_2(x, y_2)
\end{equation}

instead of ~\ref{eq:multi}.}

This substitution is motivated by the following considerations. Let $\theta \in \Theta$ represents a parameter that requires updating. For simplicity, we'll denote the partial derivative with respect to $\theta$ as $\partial_\theta$. We derive the two following gradients of $L_2$ and $\tilde{L}_2$ with respect to $\theta$:

\begin{equation}
    \begin{split}
    \partial_\theta L_2(x, y_2) &=\partial_\theta\left\{\left( y_2-f_2(x,\theta)\right)^2\right\}\\
    &= 2\partial_\theta f_2(x, \theta)\left(f_2(x, \theta) - y_2\right)
    \end{split}
\end{equation}

\begin{equation}
    \begin{split}
    \partial_\theta \tilde{L}_2(x, y_2) 
    &= \partial_\theta\left\{\left( \log(y_2+1) - \log\left(f_2(x,\theta)+1\right)\right)^2\right\}\\
    &= 2\frac{\partial_\theta f_2(x, \theta)}{f_2(x, \theta) + 1} \log\left(\frac{f_2(x, \theta)+1}{y_2+1}\right)
    \end{split}
\end{equation}

Now, the two gradients can be compared by examining their ratio.

\begin{equation}
    \left\lvert \frac{\partial_\theta \tilde{L}_2}{\partial_\theta L_2}\right\rvert = \left\lvert\frac{1}{\left(f_2(x, \theta) + 1\right)\left(f_2(x, \theta) - y_2\right)} \log\left(\frac{f_2(x, \theta)+1}{y_2+1}\right)\right\rvert    
\end{equation}

Here, the term $\frac{1}{f_2(x, \theta) - y_2} \log\left(\frac{f_2(x, \theta)+1}{y_2+1}\right)$ is contingent on the relative comparison between the prediction $f_2(x, \theta)$ and $y_2$. However, the term $\frac{1}{f_2(x, \theta) + 1}$ exhibits a dependency on the magnitude of the prediction. In simpler terms, when high numbers (outliers) are predicted, $L_2$ will significantly update the weights compared to $\tilde{L}_2$, potentially undermining the model's performance on the real data distribution. This characteristic underscores why $\tilde{L}_2$ is more effective in handling a wide range of numbers.

\section{Experiments}
This section introduces the baseline models, outlines our experimental configuration, and presents the results we achieved.

\subsection{\label{our models}Our models}
We present two models that incorporate the diverse solutions proposed in this research.

\noindent Model 1:\phantom{aaa} The architecture is exactly the one presented in \cite{lompo2025mediumsizedtransformersmodelsrelevant}. The training processus comprises the 2 following steps:

\begin{enumerate}
    \item We pre-finetune the pre-trained CamemBERT-bio augmented with the Label Embedding for Self-Attention (LESA) technique on the unannotated medical notes with the Masked Language Modeling (MLM) task.
    \item We then fine-tune the pre-finetuned model on the token classification task with the annotated notes. %As in \cite{lompo2024numerical} we also fine-tune the trained model with the blinded dataset which basically is the labelled dataset where all numbers are replaced by a keyword "nombre".
    %since we observed that it significantly improves the results.
\end{enumerate}
We kept the same architecture as in \cite{lompo2025mediumsizedtransformersmodelsrelevant}

\noindent Model 2:\phantom{aaa} We incorporated Xval architecture to Model 1, in order to have both contextual and magnitude representations of numbers. 

\subsection{Baselines}

{\color{black}
\noindent GPT-4: In these experiments, we selected a subset of 200 entries from our dataset and manually provided them to ChatGPT using a few-shot learning prompting approach. The model generated responses as sequences of numerical values, each mapped to its respective category. These outputs were then aggregated and used to compute the evaluation metrics detailed below.
}

\noindent DistillCamemBERT:\phantom{aaa}DistilCamemBERT, as described in \cite{delestre:hal-03674695}, employs Knowledge Distillation \cite{Bucila2006ModelC}, \cite{hinton2015distilling} in order to reduce the parameters of CamemBERT. This technique is a training method where a larger, expert model (teacher) guides a smaller model (student) to mimic its behavior. DistillCamemBERT has achieved performance levels comparable to CamemBERT on various NLP tasks, providing a benchmark for evaluating our approach against parameter reduction techniques. For this experiment, we utilized the version available on Huggingface \cite{wolf2019huggingface}.

\noindent Camembert-bio:\phantom{aaa} This transformer based model is a french version of BioBERT \cite{Lee_2019} introduced by \cite{touchent2023camembert} and trained on a large medical corpus.

\noindent CamemBERT-bio + LESA: This model follows the training approach detailed in \cite{lompo2025mediumsizedtransformersmodelsrelevant}, which consists in incorporating LESA to CamemBERT-bio. The authors derived two models from this approach: one model trained on the raw dataset and another where numbers are replaced with a keyword placeholder before training. We will only keep the second one as it got the best performances. We will refer to this baseline as CamemBERT-bio + LESA

\noindent CamemBERT-bio + Xval: This model follows the training approach detailed in the work by \cite{golkar2023xval}, which consists in incorporating some magnitude information into the numbers embeddings. The architecture was imported from \href{https://github.com/PolymathicAI/xVal/blob/main/xval/numformer.py}{their GitHub repository}.

\noindent NumBERT:\phantom{aaa} As intruduced in \cite{zhang2020language}, it is a version of the BERT model trained by substituting each numerical instance in the training dataset with its scientific notation representation, comprising both an exponent and mantissa. We used the implementation provided in \href{https://github.com/google-research/google-research/tree/master/numbert}{their GitHub repository}.

\noindent ELMO:\phantom{aaa} As introduced in \cite{sarzynska2021detecting}, Embeddings from Languages Models (ELMO) is a bidirectional LSTM that is trained with a coupled language model (LM) objective on a large text corpus. ELMO is very effective in practice in medical related classification tasks. We used the implementation from \href{https://github.com/HIT-SCIR/ELMoForManyLangs?tab=readme-ov-file}{their GitHub repository}.

\subsection{\label{the setup}Setup and Evaluation}
 The performance evaluation metric utilized is the $F_1$ score, a class-wise metric particularly suitable for imbalanced datasets. For the pre-finetuning step, we trained all our models for 100 epochs as a warm up with AdamW optimizer \cite{loshchilov2017decoupled} optimizer and a learning rate of $3e-5$. Then we used cosine learning scheduler until early-stopping with 4 epochs of patience. For the classification downstream task, we performed 10 sets of runs with different random seeds until early stopping. {\color{black} We compute the mean F1 score over the 10 runs along with their standard deviations, with all results summarized in Table ~\ref{results}}. For all the methods considered in this study, we used cross-entropy loss with appropriate weights for each classes in order to compensate the distribution imbalance. However, such weights were not necessary for our proposed Model 2 which performed better without them. The training, validation, and testing datasets were structured to maintain a distribution of 70\%, 15\%, and 15\% respectively across all eight classes. Our models' training process took 20 hours on a Tesla V100 GPU with 32 GB of memory. In contrast, training the baseline models required approximately 10 hours. %The entire codebase, without the dataset and the trained model weights, can be accessed on our \href{https://github.com/Aser97/HUGGINGFACE_2.git}{GitHub repository}.
 
\section{Results and Discussions}

%{\color{green}below above the y=x line}
Table ~\ref{results} shows that all models exhibit nearly perfect $F_1$ scores for the Out-of-class class. This was expected given its extensive representation. To evaluate overall performance, we compute the average $F_1$ score across all classes without applying weighting, considering that the remaining classes hold equal importance in terms of size. This choice is made to ensure a straightforward evaluation without introducing unnecessary complexity. The overall $F_1$ scores of all models are presented in Table ~\ref{results global}.

\begin{table}[H]
%\begin{subtable}
\centering
\caption{Overall $F_1$ score per model}
\begin{tabular}{|l|l|}
\hline
\textbf{Model} & \textbf{Overall} $\mathbf{F_1}$ \textbf{score} \\
\hline
\color{black} GPT-4 & \color{black} 0.95\\
\textbf{ELMO} & 0.72 \\
\textbf{DistilCamemBERT} & 0.74 \\
\textbf{Camembert-bio} & 0.69 \\
\textbf{Camembert-bio + LESA} & 0.83\\
%\textbf{Camembert-bio + LESA (2)} & 0.82 \\
\textbf{Camembert-bio + Xval} & 0.80 \\
\textbf{NumBERT} & 0.77 \\
\textbf{Model 1} & 0.83 \\
\textbf{Model 2} & \textbf{0.90} \\
\hline
\end{tabular}
\label{results global}
%\end{subtable}
\end{table}

Using this ranking as our basis, it becomes evident that our two models (models 1 and 2) consistently surpass all the other baseline methods {\color{black}except for GPT-4} 

\subsubsection{IMPACT of PRE-FINETUNING via MLM}
Table \ref{comparison} presents a performance comparison of CamemBERT-bio and CamemBERT-bio + LESA before and after prefinetuning. The results show that Prefinetuning has no significant impact on CamemBERT-bio model, which is expected since CamemBERT-bio was already trained on diverse medical datasets similar to our current dataset, which means that most of the medical knowledge was already encoded in its parameters. However, the results show a significant performance boost for CamemBERT-bio + LESA after prefinetuning. This improvement is attributed to the LESA technique and it indicates that incorporating Label Embeddings in the model's architecture requires fine-tuning in order to adjust the parameters of CamemBERT-bio's embedding layers. Essentially, the latent space derived from CamemBERT-bio + LESA evolves to better discriminate the eight categories, thereby enhancing the $F_1$ score.

\begin{table}[H]
%\begin{subtable}
\centering
\caption{Impact of prefinetuning on $F_1$ score}
\begin{tabular}{|l|l|l|}
\hline
\textbf{Model} & \textbf{Before} & \textbf{After} \\
\hline
\textbf{Camembert-bio} & 0.69 & 0.70\\
%\textbf{Camembert-bio + LESA (1)} & 0.74 & 0.83\\
\textbf{Camembert-bio + LESA} & 0.79 & 0.83 \\
\hline
\end{tabular}
\label{comparison}
%\end{subtable}
\end{table}

\subsubsection{RESULTS of MULTIPLE OBJECTIVE LOSS}: 
\begin{figure}
\centering
\includegraphics[scale=0.42]{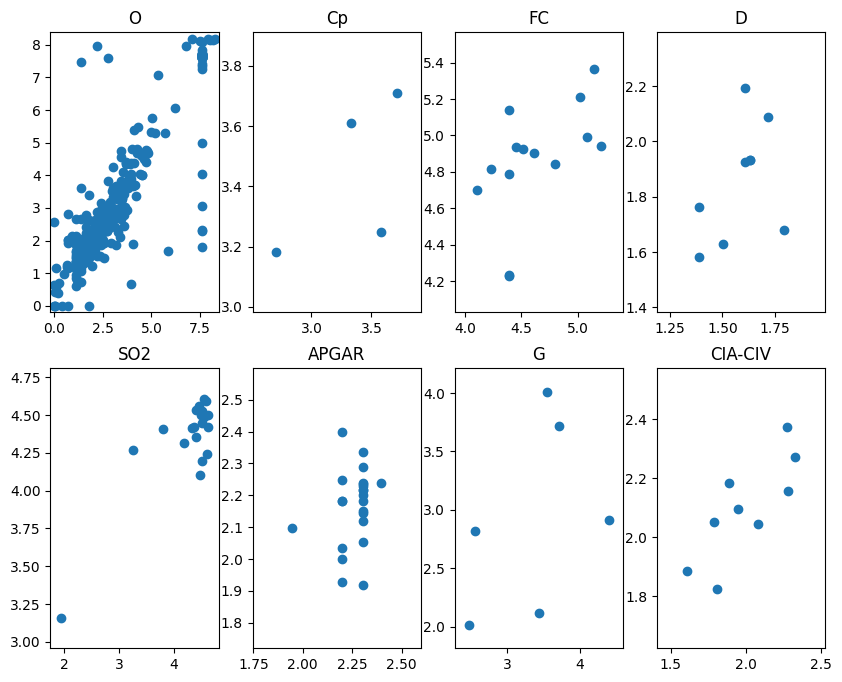}
  \caption{The numbers predicted by Model 2. The $x$ axis represents the ground truth values, and the $y$ axis represents the predicted values. Both axes are displayed on a logarithmic scale}
  \label{fig:predictions}
\end{figure}

The primary advantage of the number prediction loss function is that it introduces intrinsic magnitude information into the model's embeddings, an aspect significantly enhanced by incorporating Xval. Through this loss function, we aim for the model to reconstruct numbers with similar magnitudes to the original values. While predicting the exact numbers would lead to memorization, we expect the predictions to be close but not identical. As shown in Figure ~\ref{fig:predictions}, Model 2 the model predictions on the $y$ axis mostly aligns with the ground truth numbers on the $x$ axis. This shows that the goal was successfully achieved. We did not include the mean square error, as it is not a key metric in our evaluation. {\color{black} Since our goal is not to have predictions overly close to the ground truth, emphasizing this metric could encourage the model to memorize rather than generalize.

Table ~\ref{results} presents the F1 score performance of all studied models, broken down by category. For each category, the highest F1 score achieved by a model, excluding GPT-4, is highlighted in bold.} A detailed examination of the results in Table ~\ref{results} reveals that Model 2 outperforms all other models except GPT-4 in nearly every class. Notably, all models, including GPT-4 struggle with the \textit{ventricular Gradient} class. This difficulty arises because this class is ambiguous and can be confused with other similar classes, such as \textit{aortic gradient, aortic ring, pulmonary gradient, and gradient between the heart earcups}. This highlights a limitation of context-based token representation and explains the poor performance of the other models. {\color{black} Indeed, in certain failure cases, the surrounding context provides insufficient cues to accurately determine the gradient type. Instead, the key distinguishing factor lies in the magnitude of the gradient values. For example, gradients between heart earcups are notably lower than ventricular gradients. This magnitude-based distinction gives Model 2 a significant advantage over the other models, resulting in a higher F1 score for the gradient category.}

\begin{figure}
\centering
\includegraphics[width=1\linewidth]{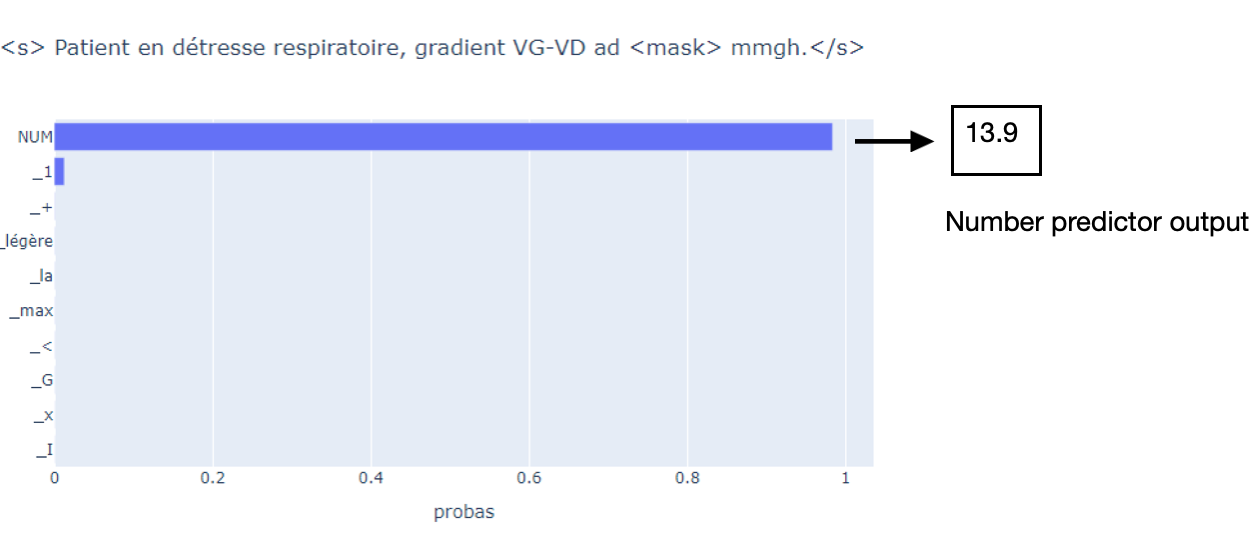}
  \caption{Output of a completion task by Model 2. The input sentence is \textit{``Patient en détresse respiratoire, gradient VG-VD ad $\langle$ mask $\rangle$ mmgh."}. The figure contains the possible words to fill the mask with their corresponding probabilities. Here the model successfully predict a number which is given by the number prediction head}
  \label{completion with Model 2}
\end{figure}

When we subject Model 2 to the same sentence completion test as in section ~\ref{tokenization problem}, the results are shown in Figure \ref{completion with Model 2}.

\begin{figure}
\centering
\includegraphics[width=1\linewidth]{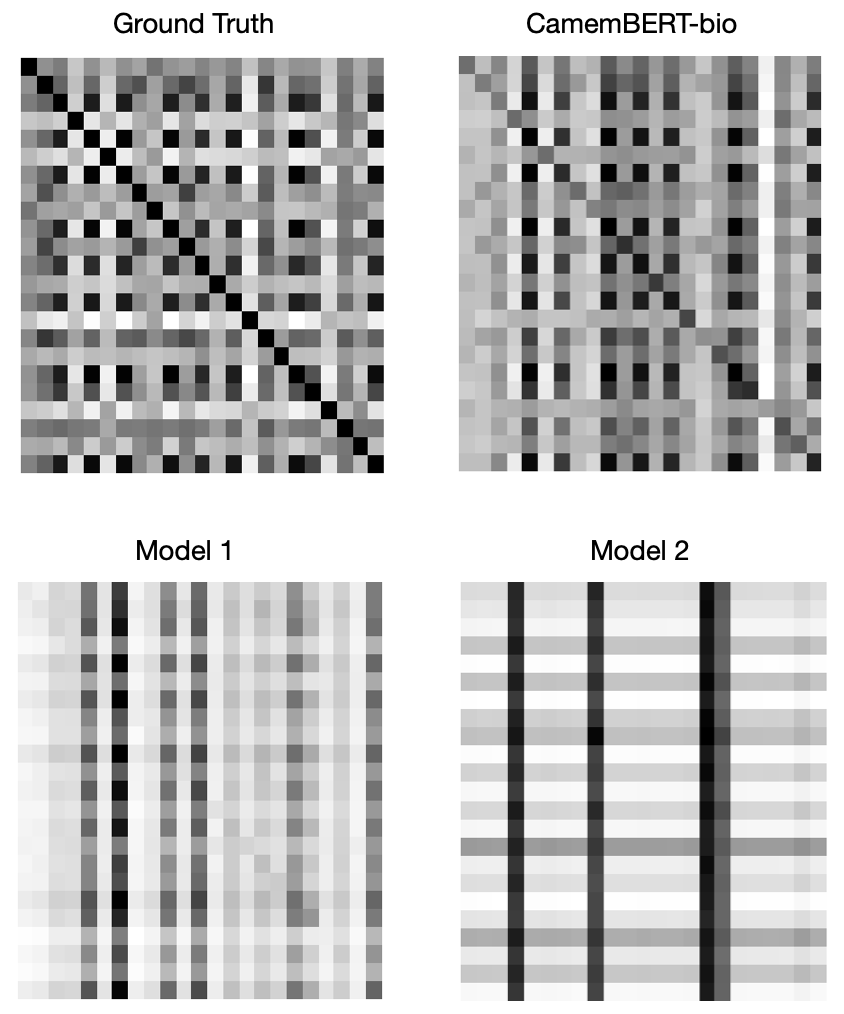}
  \caption{Evaluation of the impact of Multiple Objective Representation on the words embeddings. From left to right we computed the cosine similarity of multiple medical terms embeddings generated by CamemBERT-bio and itself, finetuned CamemBERT-bio, Model 1 and Model 2 respectively. The darker a cell is, the higher its value}
  \label{comparison with Model 2}
  \vspace{-5mm}
\end{figure}

\begin{table*}
\centering
\footnotesize
\caption{$F_1$ score results per class.}
\label{results}

\begin{subtable}
\centering
\begin{tabular}{|l|l|l|l|l|}
\hline
 & \textbf{O} &  \textbf{Cp} & \textbf{FC} & \textbf{D} \\
\hline
\color{black} GPT-4 & \color{black} 0.99& \color{black} 0.93& \color{black} 0.99 & \color{black} 0.97\\ 
\textbf{ELMO} & 0.99 $\pm$ 0.00 & 0.76 $\pm$ 0.25 & 0.62 $\pm$ 0.18 & 0.68 $\pm$ 0.13 \\
\textbf{DistilCamemBERT} & 0.99 $\pm$ 0.00 & 0.80 $\pm$ 0.16 & 0.62 $\pm$ 0.09 & 0.69 $\pm$ 0.14 \\
\textbf{Camembert-bio} & 0.99 $\pm$ 0.00 & 0.78 $\pm$ 0.27 & 0.68 $\pm$ 0.07 & 0.52 $\pm$ 0.11 \\
\textbf{CamemBERT-bio + LESA} & 0.99 $\pm$ 0.00 & 0.85 $\pm$ 0.07 & 0.87 $\pm$ 0.04 & 0.84 $\pm$ 0.13 \\
%\textbf{CamemBERT-bio + LESA (2)} & 0.99 $\pm$ 0.00 & 0.92 $\pm$ 0.08 & 0.83 $\pm$ 0.08 & 0.81 $\pm$ 0.14 \\
\textbf{CamemBERT-bio + Xval} & 0.99 $\pm$ 0.00 & 0.85 $\pm$ 0.14 & 0.64 $\pm$ 0.07 & \textbf{0.92 $\pm$ 0.06} \\
\textbf{NumBERT} & 0.99 $\pm$ 0.00 & 0.78 $\pm$ 0.10 & 0.64 $\pm$ 0.04 & 0.67 $\pm$ 0.15 \\
\textbf{Model 1} & 0.99 $\pm$ 0.00 & 0.85 $\pm$ 0.07 & 0.87 $\pm$ 0.04 & 0.84 $\pm$ 0.13 \\
\textbf{Model 2} & 0.99 $\pm$ 0.00 & \textbf{0.86 $\pm$ 0.07} & \textbf{0.89 $\pm$ 0.11} & 0.90 $\pm$ 0.08 \\
\hline
\end{tabular}
\end{subtable}
%\end{table}

%\begin{table}

\begin{subtable}
\centering
\begin{tabular}{|l|l|l|l|l|}
\hline
 & \textbf{SO2} & \textbf{AGPR} & \textbf{G} & \textbf{CIA/CIV} \\
\hline
\color{black} GPT-4 & \color{black} 0.99 & \color{black} 0.99 & \color{black} 0.78 & \color{black} 0.99 \\
\textbf{ELMO} & 0.83 $\pm$ 0.15 & 0.87 $\pm$ 0.12 & 0.24 $\pm$ 0.14 & 0.81 $\pm$ 0.16 \\
\textbf{DistilCamemBERT} & 0.84 $\pm$ 0.05 & 0.95 $\pm$ 0.02 & 0.23 $\pm$ 0.02 & 0.81 $\pm$ 0.15 \\
\textbf{Camembert-bio} & 0.85 $\pm$ 0.05 & 0.93 $\pm$ 0.02 & 0.22 $\pm$ 0.02 & 0.80 $\pm$ 0.08 \\
\textbf{CamemBERT-bio + LESA} & 0.91 $\pm$ 0.02 & 0.99 $\pm$ 0.01 & 0.27 $\pm$ 0.01 & 0.91 $\pm$ 0.05\\
%\textbf{CamemBERT-bio + LESA (2)} & 0.90 $\pm$ 0.02 & 0.95 $\pm$ 0.09 & 0.26 $\pm$ 0.04 & 0.89 $\pm$ 0.06 \\
\textbf{CamemBERT-bio + Xval} & 0.90 $\pm$ 0.03 & 0.95 $\pm$ 0.01 & 0.23 $\pm$ 0.15 & 0.91 $\pm$ 0.08 \\
\textbf{NumBERT} & 0.83 $\pm$ 0.07 & 0.90 $\pm$ 0.13 & 0.24 $\pm$ 0.10 & 0.82 $\pm$ 0.11 \\
\textbf{Model 1} & 0.91 $\pm$ 0.02 & 0.99 $\pm$ 0.01 & 0.27 $\pm$ 0.01 & 0.91 $\pm$ 0.05\\
\textbf{Model 2} & \textbf{0.93 $\pm$ 0.02} & \textbf{0.98 $\pm$ 0.03} & \textbf{0.61 $\pm$ 0.12} & \textbf{0.94 $\pm$ 0.06} \\
\hline
\end{tabular}
\end{subtable}
%\end{table}
\vspace{-5mm}
%\begin{table}
\end{table*}

{\color{black} The side-by-side comparison of Figure ~\ref{completion with camembert bio} and ~\ref{completion with Model 2} is highly illustrative. Unlike CamemBERT-bio, Model 2 clearly recognizes that a numerical value is expected in this context. Moreover, when we take the prediction from the Number Head layer, we obtain $13.9$, which falls within the correct range for this type of gradient. We attribute this improvement to the implementation of multiple objective number representations.} By replacing numbers with placeholders, the unstructured and noisy text becomes clearer, making it easier for the model to learn where numbers are expected. Furthermore, incorporating LESA and Xval enhances the model's ability to more accurately predict the value range of the masked numbers. {\color{black} We aim to illustrate our claim by measuring the similarity of the embeddings vectors produced by various models derived from CamemBERT-bio and those generated by CamemBERT-bio itself (our reference). Specifically, we evaluate three derived models: CamemBERT-bio fine-tuned on unannotated medical notes using masked language modeling (MLM), Model 1, and Model 2. This experiment is designed to assess how the different methodologies developed in this study influence the embedding space of CamemBERT-bio. To conduct this analysis, we selected twenty frequently occurring medical terms from our dataset. As a similarity metric, we employed cosine similarity, following the insights from \cite{steck2024cosine}, which highlights that cosine similarity is a meaningful measure primarily for models trained with respect to this metric—a condition that can be facilitated by layer normalization. Since our models satisfy this prerequisite, cosine similarity was chosen as the most appropriate metric for our evaluation.}

The cosine similarity graphics are displayed in Figure ~\ref{comparison with Model 2}. In this figure, each plot represents a cosine similarity matrix $M$ in grayscale. Each row $i$ and column $j$ is indexed by the medical terms $\textrm{tok}_i$ and $\textrm{tok}_j$, respectively. The corresponding cell $M_{i,j}$ contains the cosine similarity score of the embeddings of $\textrm{tok}_i$ obtained by the reference model and the embedding of $\textrm{tok}_j$ produced by the model we're comparing. The darker a cell is, the higher the cosine similarity score.

{\color{black} From left to right, the matrices progressively lighten, indicating increasing divergence of the embedding vectors from the reference model. Specifically, the second matrix demonstrates that CamemBERT-bio embeddings remain largely unchanged after fine-tuning, corroborating the results in Table ~\ref{comparison}, which suggest that fine-tuning alone does not introduce additional knowledge into the model. In contrast, the third matrix reveals a notable difference in embeddings between Model 1 and CamemBERT-bio. This observation aligns with the analysis in Table ~\ref{comparison}, which suggests that fine-tuning Model 1 via MLM modifies the embedding space due to the integration of label information into CamemBERT-bio through the LESA technique. Finally, the last matrix in Figure ~\ref{comparison with Model 2} exhibits the most substantial deviation from the reference, highlighting the impact of Multiple Objective Representations. This result supports the hypothesis that incorporating multiple losses leads to a more significant transformation of the embedding space compared to using LESA alone.}

{
\color{black}
\subsection{\label{sec:gpt vs model 2}GPT-4 VS MODEL 2}

The category-wise and overall F1 score analysis highlights GPT-4’s superior performance, with an overall F1 score surpassing Model 2 by 0.05 units. However, this improvement is relatively minor given the vast disparity in model size—Model 2, with 110 million parameters, compared to GPT-4’s multi-trillion parameter scale.  

A failure case analysis conducted by \cite{lompo2025mediumsizedtransformersmodelsrelevant} identified recurring misclassifications by GPT-4, particularly in the Contractibility (Cp) and Gradient (G) categories. Misclassifications in the Cp category often stemmed from ambiguous abbreviations, whereas errors in the G category frequently involved confusion between different gradient types, such as aortic and atrial/ventricular gradients, which were mistakenly linked to the interventricular gradient. These systematic errors suggest limitations in GPT-4’s contextual comprehension of medical terminology. Similarly, Table ~\ref{results} shows that Model 2 exhibits difficulties in the Cp and G categories. However, it significantly reduced the performance gap compared to the difference between CamemBERT-bio and GPT-4, narrowing it from 0.52 to 0.17 F1 score units in the G category.  

From a practical perspective, our approach offers notable advantages in terms of computational efficiency and system integration. Unlike large-scale LLMs such as GPT-4 and LLaMA, Model 2 can be seamlessly deployed in medical decision-support systems while maintaining competitive performance. Its reduced computational demands enable local execution, eliminating reliance on external cloud-based models and thereby enhancing data privacy and accessibility.}

{\color{black}
\section{Limitations and Application Perspectives}
The main limitation of this study is the choice of CamemBERT-bio over large language models (LLMs). In a context where scalable models are emphasized, this decision was driven by the need to balance performance with computational efficiency. While a direct comparison with GPT-4 confirms its superior overall performance, its computational demands are significantly higher, making it less practical for integration into resource-constrained environments.  

Additionally, we recognize that our experiments focus on a specific task and rely on a single dataset. However, we believe that the methodologies introduced in this work are broadly applicable and can be adapted to similar scenarios involving limited data availability.  

The Clinical Decision Support System (CDSS) at CHUSJ is still under development. Our next step is to integrate this NLP model for clinical text analysis with previously developed algorithms for heart failure detection \cite{le2022detecting, le2023adaptation, le2023improving}, hypoxemia detection \cite{sauthier2021estimated}, and chest X-ray analysis \cite{zaglam2014computer, yahyatabar2020dense}. Ultimately, we aim to deploy a comprehensive CDSS within the pediatric intensive care unit (PICU) at CHUSJ to facilitate early diagnosis of acute respiratory distress syndrome (ARDS). Once integrated into the PICU's e-Medical infrastructure, we will conduct a prospective evaluation of the system’s ARDS detection capabilities, using the results to guide further refinements and enhancements.}

\section{Conclusion}

This research focused on optimizing numerical understanding by CamemBERT-bio, particularly for classifying numerical values in small medical datasets. We addressed the challenges posed by tokenization and the lack of text structure, which required special handling of numbers. To tackle these issues, we implemented multiple approaches. First, we demonstrated that CamemBERT-bio + LESA could be further improved through prefinetuning with the MLM task, an improvement not seen with CamemBERT-bio alone. Additionally, we developed a multiple-objective numbers representation by combining LESA \cite{lompo2025mediumsizedtransformersmodelsrelevant} and Xval \cite{golkar2023xval}, resulting in embeddings that incorporate both contextual and magnitude information. Rigorous evaluations of these strategies showed significant enhancements in the performance of CamemBERT-bio. {\color{black}While our results do not surpass those of GPT-4, they remain comparable despite the significant difference in model size. This demonstrates the effectiveness of our approach in achieving a balance between performance and computational efficiency. We believe that this work presents a promising alternative to large-scale LLMs for medical note processing, particularly in hospital environments where computational resources are limited and data availability constraints must be considered.}

%\section*{Acknowledgment}
%The medical notes utilized in this study were provided by CHU Sainte Justine Hospital and annotated by Dr. Jérôme Rambaud and Dr. Guillaume Sans. The authors express their gratitude to physiotherapist Kevin Albert for his contribution to the literature review on medical benchmarks.

%This research was partially supported by the Natural Sciences and Engineering Research Council (NSERC), the Institut de Valorisation des Données de l’Université de Montréal (IVADO), and the Fonds de la Recherche en Santé du Québec (FRQS).

\flushend
\bibliographystyle{IEEEtran}
%\flushend
\bibliography{Bibliography}

\begin{thebibliography}{10}
\providecommand{\url}[1]{#1}
\csname url@rmstyle\endcsname
\providecommand{\newblock}{\relax}
\providecommand{\bibinfo}[2]{#2}
\providecommand\BIBentrySTDinterwordspacing{\spaceskip=0pt\relax}
\providecommand\BIBentryALTinterwordstretchfactor{4}
\providecommand\BIBentryALTinterwordspacing{\spaceskip=\fontdimen2\font plus
\BIBentryALTinterwordstretchfactor\fontdimen3\font minus
  \fontdimen4\font\relax}
\providecommand\BIBforeignlanguage[2]{{%
\expandafter\ifx\csname l@#1\endcsname\relax
\typeout{** WARNING: IEEEtran.bst: No hyphenation pattern has been}%
\typeout{** loaded for the language `#1'. Using the pattern for}%
\typeout{** the default language instead.}%
\else
\language=\csname l@#1\endcsname
\fi
#2}}

\bibitem{touchent2023camembert}
R.~Touchent, L.~Romary, and E.~de~La~Clergerie, ``Camembert-bio: a tasty french
  language model better for your health,'' \emph{arXiv preprint
  arXiv:2306.15550}, 2023.

\bibitem{lompo2025mediumsizedtransformersmodelsrelevant}
\BIBentryALTinterwordspacing
B.~A. Lompo, T.-D. Le, P.~Jouvet, and R.~Noumeir, ``Are medium-sized
  transformers models still relevant for medical records processing?'' 2025.
  [Online]. Available: \url{https://arxiv.org/abs/2404.10171}
\BIBentrySTDinterwordspacing

\bibitem{golkar2023xval}
S.~Golkar, M.~Pettee, M.~Eickenberg, A.~Bietti, M.~Cranmer, G.~Krawezik,
  F.~Lanusse, M.~McCabe, R.~Ohana, L.~Parker, \emph{et~al.}, ``xval: A
  continuous number encoding for large language models,'' \emph{arXiv preprint
  arXiv:2310.02989}, 2023.

\bibitem{wallace2019nlp}
E.~Wallace, Y.~Wang, S.~Li, S.~Singh, and M.~Gardner, ``Do nlp models know
  numbers? probing numeracy in embeddings,'' \emph{arXiv preprint
  arXiv:1909.07940}, 2019.

\bibitem{cui2019regular}
M.~Cui, R.~Bai, Z.~Lu, X.~Li, U.~Aickelin, and P.~Ge, ``Regular expression
  based medical text classification using constructive heuristic approach,''
  \emph{IEEE Access}, vol.~7, pp. 147\,892--147\,904, 2019.

\bibitem{chen2023improving}
C.-C. Chen, H.~Takamura, I.~Kobayashi, and Y.~Miyao, ``Improving numeracy by
  input reframing and quantitative pre-finetuning task,'' in \emph{Findings of
  the Association for Computational Linguistics: EACL 2023}, 2023, pp. 69--77.

\bibitem{charton2021linear}
F.~Charton, ``Linear algebra with transformers,'' \emph{arXiv preprint
  arXiv:2112.01898}, 2021.

\bibitem{Loukas_2022}
\BIBentryALTinterwordspacing
L.~Loukas, M.~Fergadiotis, I.~Chalkidis, E.~Spyropoulou, P.~Malakasiotis,
  I.~Androutsopoulos, and G.~Paliouras, ``Finer: Financial numeric entity
  recognition for xbrl tagging,'' in \emph{Proceedings of the 60th Annual
  Meeting of the Association for Computational Linguistics (Volume 1: Long
  Papers)}.\hskip 1em plus 0.5em minus 0.4em\relax Association for
  Computational Linguistics, 2022. [Online]. Available:
  \url{http://dx.doi.org/10.18653/v1/2022.acl-long.303}
\BIBentrySTDinterwordspacing

\bibitem{Nguyen2018EmbNumSL}
\BIBentryALTinterwordspacing
P.~Nguyen, K.~Nguyen, R.~Ichise, and H.~Takeda, ``Embnum: Semantic labeling for
  numerical values with deep metric learning,'' \emph{ArXiv}, vol.
  abs/1807.01367, 2018. [Online]. Available:
  \url{https://api.semanticscholar.org/CorpusID:49567815}
\BIBentrySTDinterwordspacing

\bibitem{cheng2024potential}
N.~Cheng, Z.~Yan, Z.~Wang, Z.~Li, J.~Yu, Z.~Zheng, K.~Tu, J.~Xu, and W.~Han,
  ``Potential and limitations of llms in capturing structured semantics: A case
  study on srl,'' in \emph{International Conference on Intelligent
  Computing}.\hskip 1em plus 0.5em minus 0.4em\relax Springer, 2024, pp.
  50--61.

\bibitem{shah2024accuracy}
S.~V. Shah, ``Accuracy, consistency, and hallucination of large language models
  when analyzing unstructured clinical notes in electronic medical records,''
  \emph{JAMA Network Open}, vol.~7, no.~8, pp. e2\,425\,953--e2\,425\,953,
  2024.

\bibitem{ullah2024challenges}
E.~Ullah, A.~Parwani, M.~M. Baig, and R.~Singh, ``Challenges and barriers of
  using large language models (llm) such as chatgpt for diagnostic medicine
  with a focus on digital pathology--a recent scoping review,''
  \emph{Diagnostic pathology}, vol.~19, no.~1, p.~43, 2024.

\bibitem{Li2018}
\BIBentryALTinterwordspacing
Y.~Li, L.~Yao, C.~Mao, A.~Srivastava, X.~Jiang, and Y.~Luo, ``Early prediction
  of acute kidney injury in critical care setting using clinical notes,'' in
  \emph{2018 {IEEE} International Conference on Bioinformatics and Biomedicine
  ({BIBM})}.\hskip 1em plus 0.5em minus 0.4em\relax {IEEE}, Dec. 2018.
  [Online]. Available: \url{https://doi.org/10.1109/bibm.2018.8621574}
\BIBentrySTDinterwordspacing

\bibitem{ezen2020comparison}
A.~Ezen-Can, ``A comparison of lstm and bert for small corpus,'' \emph{arXiv
  preprint arXiv:2009.05451}, 2020.

\bibitem{mascio2020comparative}
A.~Mascio, Z.~Kraljevic, D.~Bean, R.~Dobson, R.~Stewart, R.~Bendayan, and
  A.~Roberts, ``Comparative analysis of text classification approaches in
  electronic health records,'' \emph{arXiv preprint arXiv:2005.06624}, 2020.

\bibitem{Bucila2006ModelC}
C.~Bucila, R.~Caruana, and A.~Niculescu-Mizil, ``Model compression,'' in
  \emph{Knowledge Discovery and Data Mining}, 2006.

\bibitem{Sanh2019}
\BIBentryALTinterwordspacing
V.~Sanh, L.~Debut, J.~Chaumond, and T.~Wolf, ``Distilbert, a distilled version
  of bert: smaller, faster, cheaper and lighter,'' 2019. [Online]. Available:
  \url{https://arxiv.org/abs/1910.01108}
\BIBentrySTDinterwordspacing

\bibitem{delestre2022distilcamembert}
C.~Delestre and A.~Amar, ``Distilcamembert: a distillation of the french model
  camembert,'' \emph{arXiv preprint arXiv:2205.11111}, 2022.

\bibitem{kendall2018multi}
A.~Kendall, Y.~Gal, and R.~Cipolla, ``Multi-task learning using uncertainty to
  weigh losses for scene geometry and semantics,'' in \emph{Proceedings of the
  IEEE conference on computer vision and pattern recognition}, 2018, pp.
  7482--7491.

\bibitem{wu2016google}
Y.~Wu, M.~Schuster, Z.~Chen, Q.~V. Le, M.~Norouzi, W.~Macherey, M.~Krikun,
  Y.~Cao, Q.~Gao, K.~Macherey, \emph{et~al.}, ``Google's neural machine
  translation system: Bridging the gap between human and machine translation,''
  \emph{arXiv preprint arXiv:1609.08144}, 2016.

\bibitem{kudo-richardson-2018-sentencepiece}
\BIBentryALTinterwordspacing
T.~Kudo and J.~Richardson, ``{S}entence{P}iece: A simple and language
  independent subword tokenizer and detokenizer for neural text processing,''
  in \emph{Proceedings of the 2018 Conference on Empirical Methods in Natural
  Language Processing: System Demonstrations}.\hskip 1em plus 0.5em minus
  0.4em\relax Brussels, Belgium: Association for Computational Linguistics,
  Nov. 2018, pp. 66--71. [Online]. Available:
  \url{https://aclanthology.org/D18-2012}
\BIBentrySTDinterwordspacing

\bibitem{clark2019does}
K.~Clark, U.~Khandelwal, O.~Levy, and C.~D. Manning, ``What does bert look at?
  an analysis of bert's attention,'' \emph{arXiv preprint arXiv:1906.04341},
  2019.

\bibitem{hu2024improving}
Y.~Hu, Q.~Chen, J.~Du, X.~Peng, V.~K. Keloth, X.~Zuo, Y.~Zhou, Z.~Li, X.~Jiang,
  Z.~Lu, \emph{et~al.}, ``Improving large language models for clinical named
  entity recognition via prompt engineering,'' \emph{Journal of the American
  Medical Informatics Association}, vol.~31, no.~9, pp. 1812--1820, 2024.

\bibitem{collobert2008unified}
R.~Collobert and J.~Weston, ``A unified architecture for natural language
  processing: Deep neural networks with multitask learning,'' in
  \emph{Proceedings of the 25th international conference on Machine learning},
  2008, pp. 160--167.

\bibitem{delestre:hal-03674695}
\BIBentryALTinterwordspacing
C.~Delestre and A.~Amar, ``{DistilCamemBERT : une distillation du mod{\`e}le
  fran{\c c}ais CamemBERT},'' in \emph{{CAp (Conf{\'e}rence sur l'Apprentissage
  automatique)}}, Vannes, France, July 2022. [Online]. Available:
  \url{https://hal.archives-ouvertes.fr/hal-03674695}
\BIBentrySTDinterwordspacing

\bibitem{hinton2015distilling}
G.~Hinton, O.~Vinyals, and J.~Dean, ``Distilling the knowledge in a neural
  network,'' \emph{arXiv preprint arXiv:1503.02531}, 2015.

\bibitem{wolf2019huggingface}
T.~Wolf, L.~Debut, V.~Sanh, J.~Chaumond, C.~Delangue, A.~Moi, P.~Cistac,
  T.~Rault, R.~Louf, M.~Funtowicz, \emph{et~al.}, ``Huggingface's transformers:
  State-of-the-art natural language processing,'' \emph{arXiv preprint
  arXiv:1910.03771}, 2019.

\bibitem{Lee_2019}
\BIBentryALTinterwordspacing
J.~Lee, W.~Yoon, S.~Kim, D.~Kim, S.~Kim, C.~H. So, and J.~Kang, ``{BioBERT}: a
  pre-trained biomedical language representation model for biomedical text
  mining,'' \emph{Bioinformatics}, vol.~36, no.~4, pp. 1234--1240, sep 2019.
  [Online]. Available: \url{https://doi.org/10.1093%2Fbioinformatics%2Fbtz682}
\BIBentrySTDinterwordspacing

\bibitem{zhang2020language}
X.~Zhang, D.~Ramachandran, I.~Tenney, Y.~Elazar, and D.~Roth, ``Do language
  embeddings capture scales?'' \emph{arXiv preprint arXiv:2010.05345}, 2020.

\bibitem{sarzynska2021detecting}
J.~Sarzynska-Wawer, A.~Wawer, A.~Pawlak, J.~Szymanowska, I.~Stefaniak,
  M.~Jarkiewicz, and L.~Okruszek, ``Detecting formal thought disorder by deep
  contextualized word representations,'' \emph{Psychiatry Research}, vol. 304,
  p. 114135, 2021.

\bibitem{loshchilov2017decoupled}
I.~Loshchilov and F.~Hutter, ``Decoupled weight decay regularization,''
  \emph{arXiv preprint arXiv:1711.05101}, 2017.

\bibitem{steck2024cosine}
H.~Steck, C.~Ekanadham, and N.~Kallus, ``Is cosine-similarity of embeddings
  really about similarity?'' in \emph{Companion Proceedings of the ACM on Web
  Conference 2024}, 2024, pp. 887--890.

\bibitem{le2022detecting}
T.-D. Le, R.~Noumeir, J.~Rambaud, G.~Sans, and P.~Jouvet, ``Detecting of a
  patient's condition from clinical narratives using natural language
  representation,'' \emph{IEEE open journal of engineering in medicine and
  biology}, vol.~3, pp. 142--149, 2022.

\bibitem{le2023adaptation}
------, ``Adaptation of autoencoder for sparsity reduction from clinical notes
  representation learning,'' \emph{IEEE Journal of Translational Engineering in
  Health and Medicine}, vol.~11, pp. 469--478, 2023.

\bibitem{le2023improving}
T.-D. Le, P.~Jouvet, and R.~Noumeir, ``Improving transformer performance for
  french clinical notes classification using mixture of experts on a limited
  dataset,'' \emph{arXiv preprint arXiv:2303.12892}, 2023.

\bibitem{sauthier2021estimated}
M.~Sauthier, G.~Tuli, P.~A. Jouvet, J.~S. Brownstein, and A.~G. Randolph,
  ``Estimated pao2: A continuous and noninvasive method to estimate pao2 and
  oxygenation index,'' \emph{Critical care explorations}, vol.~3, no.~10, p.
  e0546, 2021.

\bibitem{zaglam2014computer}
N.~Zaglam, P.~Jouvet, O.~Flechelles, G.~Emeriaud, and F.~Cheriet,
  ``Computer-aided diagnosis system for the acute respiratory distress syndrome
  from chest radiographs,'' \emph{Computers in biology and medicine}, vol.~52,
  pp. 41--48, 2014.

\bibitem{yahyatabar2020dense}
M.~Yahyatabar, P.~Jouvet, and F.~Cheriet, ``Dense-unet: a light model for lung
  fields segmentation in chest x-ray images,'' in \emph{2020 42nd Annual
  International Conference of the IEEE Engineering in Medicine \& Biology
  Society (EMBC)}.\hskip 1em plus 0.5em minus 0.4em\relax IEEE, 2020, pp.
  1242--1245.

\end{thebibliography}

\end{document}